\documentclass{article}

\PassOptionsToPackage{numbers, compress}{natbib}
\usepackage[final]{neurips_2024}

\usepackage{placeins} 

\usepackage[utf8]{inputenc} 
\usepackage[T1]{fontenc}    
\usepackage{hyperref}       
\usepackage{url}            
\usepackage{booktabs}       
\usepackage{amsfonts}       
\usepackage{nicefrac}       
\usepackage{microtype}      
\usepackage{xcolor}         
\usepackage{graphicx}
\usepackage{amsmath} 
\usepackage{amssymb}
\usepackage{float} 

\newcommand{\ms}{\boldsymbol{\eta}}
\newcommand{\startms}{\ms_{\varnothing}}
\newcommand{\one}{\boldsymbol{1}}
\newcommand{\stationary}{\boldsymbol{\pi}}

\title{Transformers Represent Belief State Geometry in their Residual Stream}

\author{%
Adam S. Shai\\
Simplex\\
PIBBSS\thanks{Principles of Intelligent Behaviour in Biological and Social Systems}
\And
Sarah E. Marzen\\
Department of Natural Sciences\\
Pitzer and Scripps College
\And
Lucas Teixeira\\
PIBBSS
\And
Alexander Gietelink Oldenziel\\
University College London\\
Timaeus
\And
Paul M. Riechers\\
Simplex\\
BITS
}

\begin{document}

\maketitle

\begin{abstract}
What computational structure are we building into large language models when we train them on next-token prediction? Here, we present evidence that this structure is given by the meta-dynamics of belief updating over hidden states of the data-generating process. Leveraging the theory of optimal prediction, we anticipate and then find that belief states are linearly represented in the residual stream of transformers, even in cases where the predicted belief state geometry has highly nontrivial fractal structure. We investigate cases where the belief state geometry is represented in the final residual stream or distributed across the residual streams of multiple layers, providing a framework to explain these observations. Furthermore we demonstrate that the inferred belief states contain information about the entire future, beyond the local next-token prediction that the transformers are explicitly trained on. Our work provides a general framework connecting the structure of training data to the geometric structure of activations inside transformers.
\end{abstract}

\section{Introduction}

In this work, we present a rigorous and concrete theoretical framework that connects the structure of training data to the geometry of activations in trained transformer neural networks. Our framework is grounded in the theory of optimal prediction. It suggests that transformers pretrained on next-token prediction will develop internal structures characterized by the meta-dynamics of belief updating over hidden states of the data-generating process. In other words, pretrained models should learn \emph{more} than 
the hidden structure of the data generating process---they must also learn how to update their beliefs about the hidden state of the world as they synchronize to it in context.

\begin{figure}[t]
    \centering
    \includegraphics[width=0.5\linewidth]{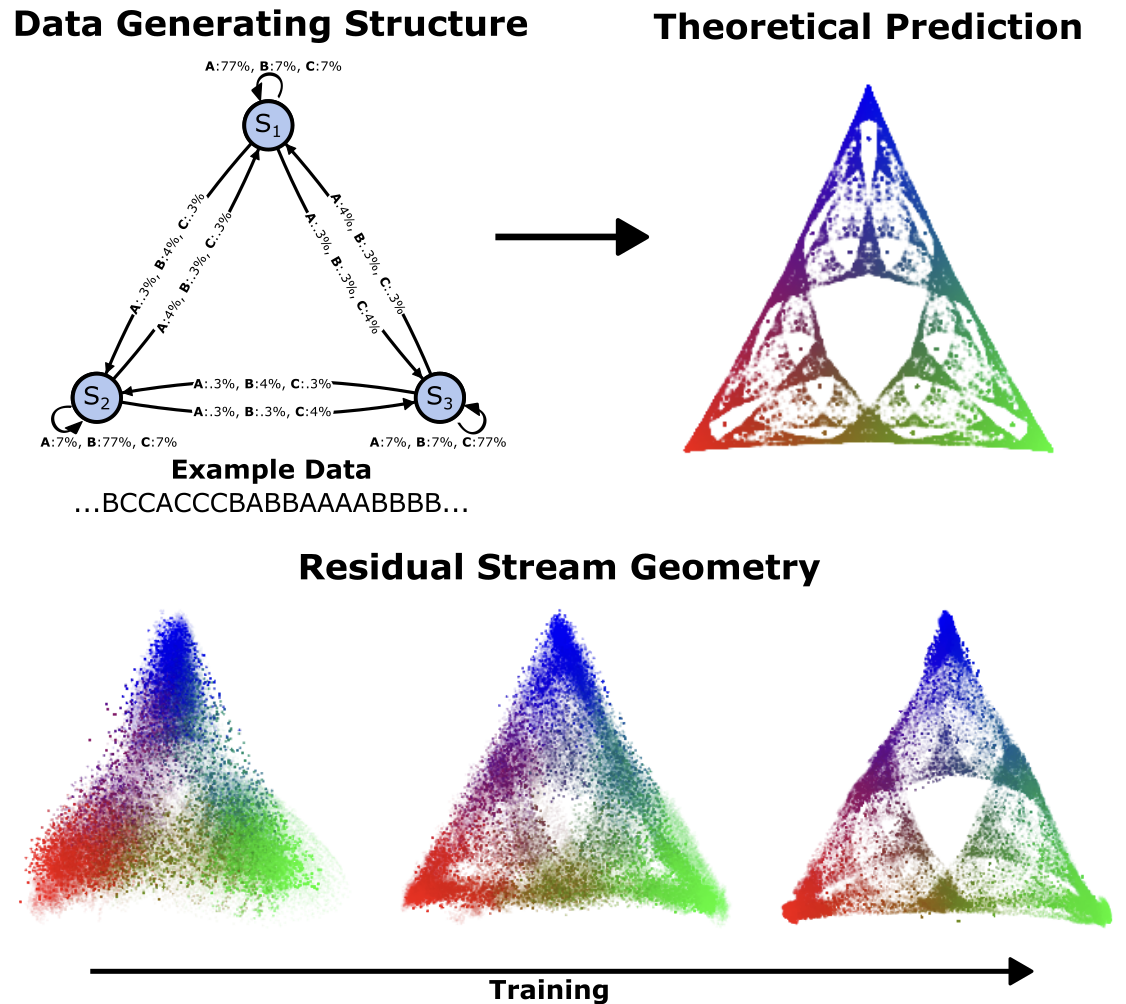}
    \caption{(Top) Given a hidden data-generating structure, our framework 
    predicts a unique
    belief state geometry in a probability simplex. Often these have highly nontrivial fractal structure as shown in this example. (Bottom) Our main experimental result is that we find that the fractal geometry of optimal beliefs is linearly embedded in the residual stream, and emerges over the course of training.}
    \label{fig:results-overview}
\end{figure}

To test this framework, we conduct well-controlled experiments where we train transformers on data generated from processes with hidden ground truth structure, and then use our theory to make predictions about the geometry of internal activations. Even in cases where the framework predicts highly nontrivial fractal structure, our empirical results confirm these predictions (Figure~\ref{fig:results-overview}). This provides a comprehensive explanation of how transformers encode information beyond the local next-token predictions they are explicitly trained on.

In Section \ref{sec:TheoryAndMethods}, we review the relevant theory from computational mechanics which motivates our prediction of the geometry of activations in the residual stream.  Core to our work is the \textit{mixed-state presentation}, which describes the metadynamic of beliefs in the probability simplex over states of the data generating process.  This formalism leads to a geometric structure that forms a natural hypothesis for the internal states of neural networks trained on prediction tasks.

In Section \ref{sec:Results}, we verify that this geometry is linearly represented in the residual stream of transformers. We implement two main experiments that control for different aspects of the hidden structure of the data generating process and make concrete predictions about the internal states of networks trained on these datasets. 
In some cases the geometry of belief state updating is found in the final residual stream, while in others it is spread out across multiple layers. We detail how our framework explains this phenomenon. After demonstrating the initial success of this framework, Section \ref{sec:Discussion} discusses the theory and implications in more depth.

\section{Theory and methods}
\label{sec:TheoryAndMethods}

\subsection{Data generating processes}

In this study, we assume that our training data is generated by an edge-emitting hidden Markov model (HMM)\footnote{For arbitrarily large, possibly non-ergodic HMMs with arbitrary initial distribution over the latent states, this does not limit the sophistication of training data.}. Paths through the HMM produce sequences of tokens from a predefined vocabulary. Tokens (\( x \in \mathcal{X} \)) are emitted as we transition between hidden states (\( s \in \mathcal{S} \)). These transitions are governed by token-labeled transition matrices \( \{ T^{(x)} \} \), where each \( T^{(x)}_{i,j} = \Pr(x, s_j | s_i) \) represents the joint probability of emitting token \( x \) 
and transitioning to state \( s_j \), given that the HMM was in state \( s_i \).

\begin{figure} [tb]
    \centering
    \includegraphics[width=0.5\linewidth]{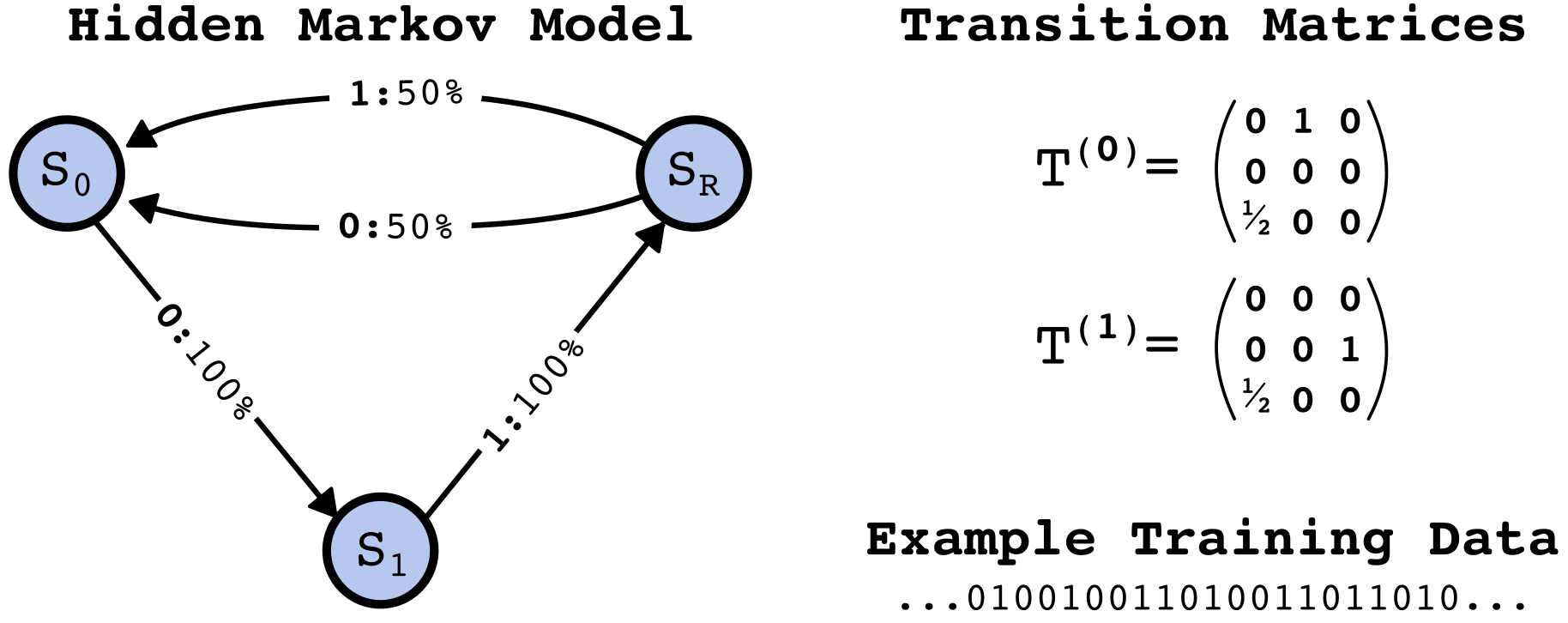}
    \caption{An illustration of a hidden Markov model (HMM) and its components. The left side shows the HMM with states 
    $\texttt{S}_\texttt{0}$,
    $\texttt{S}_\texttt{1}$, and 
    $\texttt{S}_\texttt{R}$,
    and their respective transition probabilities. The right side displays the transition matrices \( T^{(\texttt{0})} \) and \( T^{(\texttt{1})} \) corresponding to token emissions \( \texttt{0} \) and \( \texttt{1} \). Example training data is provided at the bottom, demonstrating a sequence generated by the HMM.}
    \label{fig:Z1Rexample}
\end{figure}

As a simple example, consider the HMM shown in Figure \ref{fig:Z1Rexample}, that we call the Zero-One-Random (Z1R) process. The Z1R process generates strings of tokens of the form \texttt{...01R01R...}, where \texttt{R} is a randomly generated \texttt{0} or \texttt{1}. This HMM has 3 states: 
$\texttt{S}_\texttt{0}$,
$\texttt{S}_\texttt{1}$, and 
$\texttt{S}_\texttt{R}$.
Arrows of the form 
\( s \xrightarrow{x: p\%} s' \) denote the probability
\( \Pr(x, s' | s) = p\% \) 
of moving to state \( s' \) and emitting the token \( x \), given that the process was in state \( s \).

\subsection{Mixed-state presentations}

There are competing intuitions for what transformers should represent.
Are they stochastic parrots~\cite{bender2021dangers}?
Do they build a world model~\cite{ha2018world}?
One natural intuition would be that transformers represent the hidden structure of the data-generating process---a world model---in order to predict the next token well.
For instance, Ilya Sutskever has said: "Predicting the next token well means that you understand the underlying reality that led to the creation of that token~\cite{Sutskever2023}." This type of intuition is natural, but 
not formal.

 Computational mechanics is a formalism that was developed in order to study the limits of prediction in chaotic systems, and has since expanded to a deep and rigorous theory of computational structure for any process~\cite{Crut12a}. One of its contributions is in providing a rigorous answer to what structures are necessary to perform optimal prediction~\cite{Shal98a}.
 
 Interestingly, computational mechanics shows that prediction is substantially more complicated than generation~\cite{lohr2012predictive, ruebeck2018prediction}. Informally, the reason for this is that even if an agent knows the underlying hidden generative structure that creates the data they are trying to predict, the agent must still perform computational work in order to infer which state the generative process is in, given finite observations of data. Thus, there are conceptually two distinct inference processes related to prediction: learning the hidden structure of the data generating process, and subsequently inferring which state the data generating process is in given finite observations of data.
 
\begin{figure} [t]
    \centering
    \includegraphics[width=0.98\linewidth]{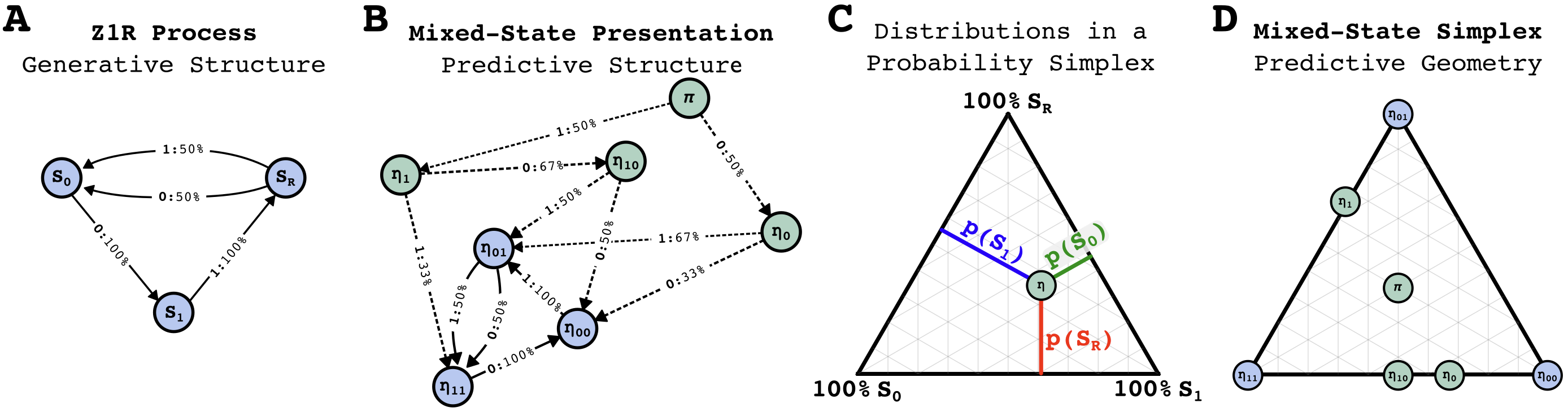}
    \caption{(A) An example generative structure called the zero-one-random process (Z1R), since it generates data of the form 
    \texttt{...01R01R01R...}
    where \texttt{R} is a random bit. (B) The generative structure implies a unique metadynamic over belief states as a predictor synchronizes to the hidden state of the world as it observes more context. This predictive structure is called the mixed-state presentation (MSP). We label belief states with $\ms_w$ where $w$ is the shortest string of emissions which leads to that belief state. (C) Belief states are distributions over generator states and can be embedded in a probability simplex. To read off this distribution for a given belief state, $\ms$, one measures the perpendicular distance from the point to each edge of the simplex, shown as blue, green, and red lines in this example. These distances directly give the probabilities for each state. Thus, the vertices represent states of certainty over one generator state, since the perpendicular distance is nonzero to only one of the edges of the simplex. (D) Plotting the belief state distributions in the probability simplex gives the belief state geometry.}
    \label{fig:Z1RMSP}
\end{figure}

Computational mechanics formally captures the computational structure of the latter inference process with the \textit{mixed-state presentation} (MSP). Whereas the HMM of a data generating process describes a generative structure (Figure~\ref{fig:Z1RMSP}A), the MSP is the structure of prediction (Figure~\ref{fig:Z1RMSP}B). The MSP answers the question of how an optimal observer updates their beliefs over the states of the data-generating process given finite observations of tokens~\citep{Riec18_SSAC1, Jurg21_Shannon}. If the observer is in a belief state given by a probability distribution \( \ms \) (a row vector) over the hidden states of the data generating process, then the update rule for the new belief state \( \ms’ \) given that the observer sees a new token $x$ is:
\begin{align}
\ms’ = \frac{\ms T^{(x)}}{\ms T^{(x)} \one}
\label{eq:ms_update}
\end{align}
where $\one$ is a column vector of ones of appropriate dimension, with the denominator ensuring proper normalization of the updated belief state. In general, starting from the initial belief state $\startms$, we can find the belief state after observing a sequence of tokens $x_0, x_1, \dots , x_N$:
\begin{align}
\ms = \frac{\startms T^{(x_0)} T^{(x_1)} \cdots T^{(x_N)}}{\startms T^{(x_0)} T^{(x_1)} \cdots T^{(x_N)} \one} ~.
\end{align}

This process of belief updating is itself another HMM, where hidden states are associated with belief states, and paths leading to particular belief states are the observed token sequence. The probability of transitioning from belief state $\ms$ to $\ms' = \ms T^{(x)} / \ms T^{(x)} \one$ of Eq.~\eqref{eq:ms_update} is simply given by the probability $\ms T^{(x)} \one$ of observing an $x$ from $\ms$. For stationary processes, the optimal initial belief state is given by the stationary distribution $\startms = \stationary$ over latent states (the left-eigenvector of the transition matrix $T=\sum_x T^{(x)}$ associated with the eigenvalue of 1). 

Each belief state of the MSP is a probability distribution over the states of the data generating process. 
Belief states thus inherit a natural geometry;
we can plot these belief states in a probability simplex, as indicated in Figure~\ref{fig:Z1RMSP}C. We refer to the geometric arrangement of the points on the simplex as the \textit{belief state geometry}. Note that each belief state---as a distribution over generator states---induces a probability density over all possible futures.
Distributions over the entire future clearly carry more than just next-token information. We can see this in our example in Figure \ref{fig:Z1RMSP}D. The states $\stationary$, $\ms_\texttt{10}$, and $\ms_\texttt{01}$ all have the same next-token prediction despite being distinct belief states.

\subsection{Finding the belief simplex in the residual stream}

\begin{figure} [tb]
    \centering
    \includegraphics[width=1\linewidth]{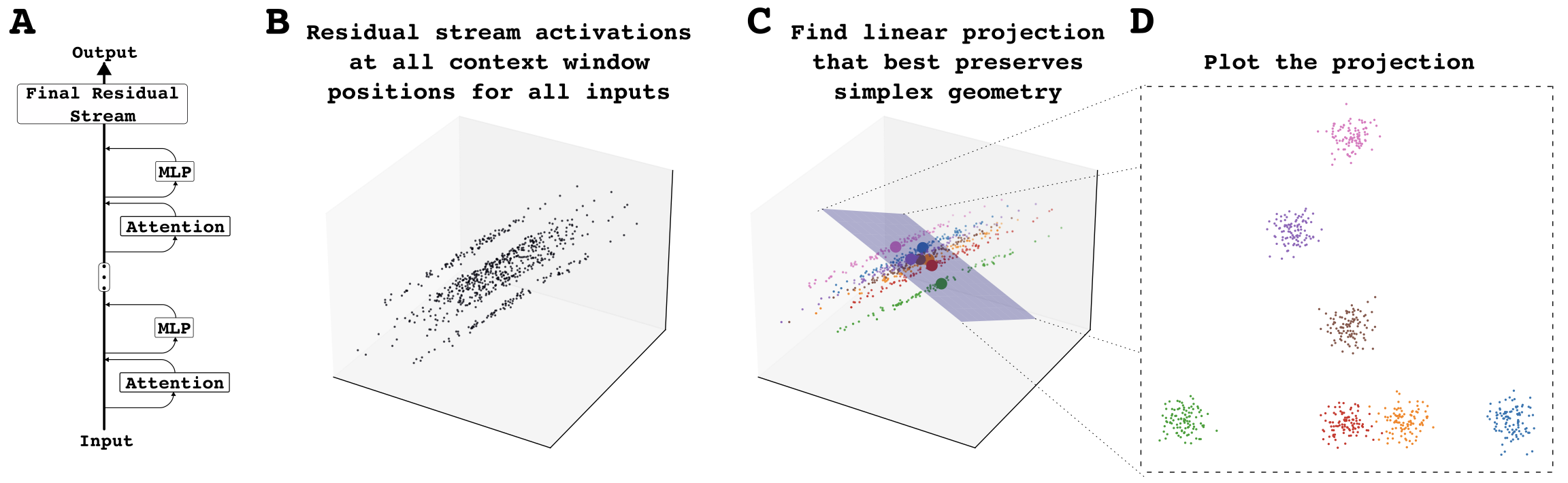}
    \caption{To verify if transformers represent belief state geometries in their residual streams, we record (A) residual stream activations at all context window positions over all inputs. (B) These activations live in a high dimensional space. (C) Each input has a ground-truth optimal belief state, which is a probability distribution over states of the data-generating process. In this way we can label, or color, each activation by the ground-truth belief associated with the input. (D) Using linear regression we then find a linear subspace of the activation space that best preserves the belief state geometry of the simplex. }
    \label{fig:procedurel}
\end{figure}

As described in the previous section, the MSP and belief state geometry are formalizations of the computational structure of the belief updating process an observer must perform in the service of optimal prediction. Consequently, a natural hypothesis would be that transformers, trained on data generated from a given ground-truth HMM, will represent the MSP structure internally. Our procedure for finding the belief states in the residual stream of a pretrained transformer 
is depicted schematically in Figure~\ref{fig:procedurel}.

We begin by training a transformer on data generated by a ground-truth HMM. We consider activations of the residual stream (in either a single layer, or a concatenation of layers) induced by all possible input sequences (Figure~\ref{fig:procedurel}AB), and at all context window positions. Thus our dataset consists of a set of activations, $a \in \mathbb{R}^{d_\text{resid}}$, when we consider a single layer at a given position, where $d_\text{resid}$ is the residual stream dimension. For each input sequence, our framework tells us which belief state the input corresponds to, and the probability distribution over hidden states of the generative process that corresponds to this belief state. These probability distributions can be thought of as points \( b \in \mathbb{R}^{|\mathcal{S}|} \), where \( |\mathcal{S}| \) denotes the number of hidden states in the generative process. 

 Since every input has a belief state associated with it\footnote{In general, multiple inputs will lead to the same belief state.}, we can label (or color, as in Figure~\ref{fig:procedurel}C) each activation by the associated ground-truth belief state. We then use standard linear regression to find an affine map from \( a \) to \( b \), of the form $b \approx Wa + c$, where \( W \in \mathbb{R}^{|\mathcal{S}| \times d_{\text{resid}}} \) is a weight matrix and \( c \in \mathbb{R}^{|\mathcal{S}|} \) is a bias vector. Parameters \( W \) and \( c \) are found by minimizing the mean squared error between the predicted belief states and the true belief states. The matrix $W$ 
 projects
 from the residual stream to (as best as possible) the probability simplex, shown diagrammatically in Figure~\ref{fig:procedurel}CD.

\section{Results}
\label{sec:Results}

\subsection{Belief state geometry is linearly represented in the residual stream}

\begin{figure} [tb]
    \centering
    \includegraphics[width=0.75\linewidth]{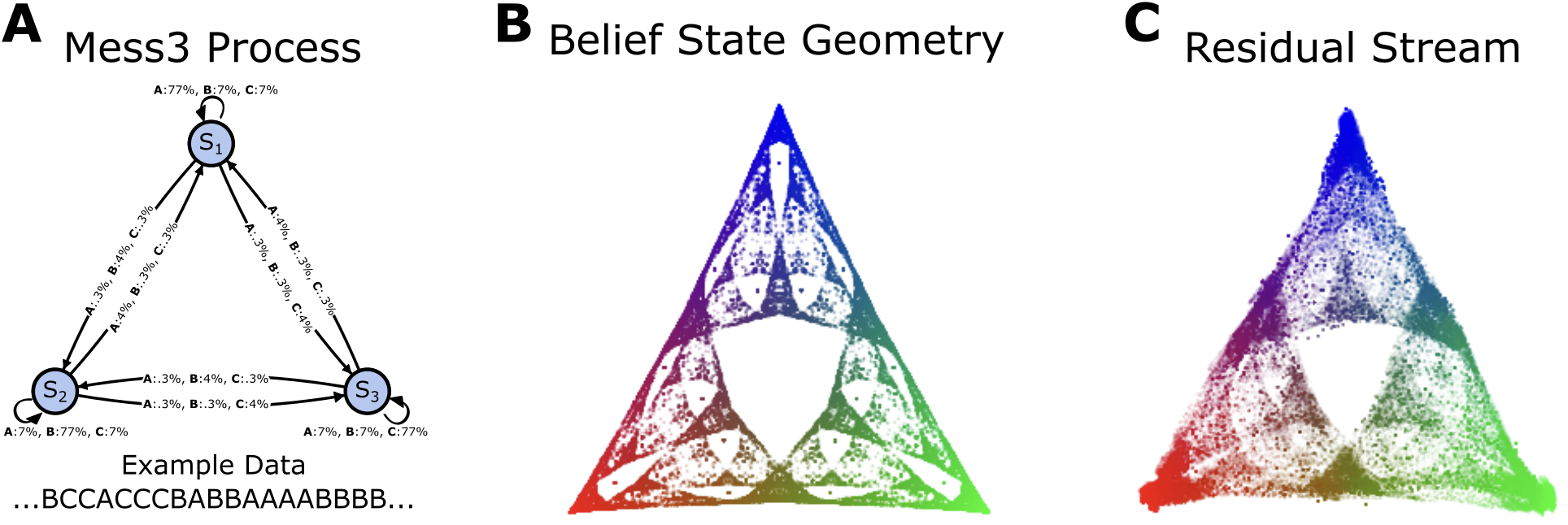}
    \caption{The residual stream of trained transformers linearly represents the belief state geometry of the mixed-state presentation. (A) The Mess3 Process has 3 hidden states and generates sequences in a token vocabulary of $\{\texttt{A}, \texttt{B},\texttt{C}\}$. (B) The ground truth belief state geometry of the Mess3 Process has intricate fractal structure. Each point in this plot is a belief state---a probability distribution over the hidden states of the Mess3 Process. Points are colored by taking the belief probability distribution and using them as RGB values. (C) We find a linear projection of the final residual stream activations contains a representation of the ground-truth belief geometry. Points are colored according to the ground-truth belief states.}
    \label{fig:mess3results}
\end{figure}

To investigate the computational structure learned by transformers we conducted an experiment using a simple 3-state hidden Markov model (HMM) called the Mess3 Process~\citep{Marz17a} (Figure \ref{fig:mess3results}). 
We used the data generated by this process (Figure \ref{fig:mess3results}) to train a transformer model.

The Mess3 MSP has a fractal structure, shown in Figure \ref{fig:mess3results}B, providing a highly non-trivial test of our theory's prediction. Each point in this geometry corresponds to a probability distribution over the hidden states of the data generating process, and thus lies in a 2-simplex.

To test these predictions, we analyzed the final layer of the transformer's residual stream, before the layer norm and unembedding. Using linear regression, we identified a 2D subspace of the 64-dimensional residual activations that best matched the ground-truth belief distributions from the MSP. Remarkably, the geometry of this 2D subspace closely resembled the predicted fractal structure (Figure~\ref{fig:mess3results}C), providing strong evidence that the transformer had indeed learned to represent the geometry of the belief states in its residual stream\footnote{These results hold when analyzing the activations after the final layer norm as well, see Figure \ref{fig:layernorm}.}.

\begin{figure} [tb]
    \centering
    \includegraphics[width=0.75\linewidth]{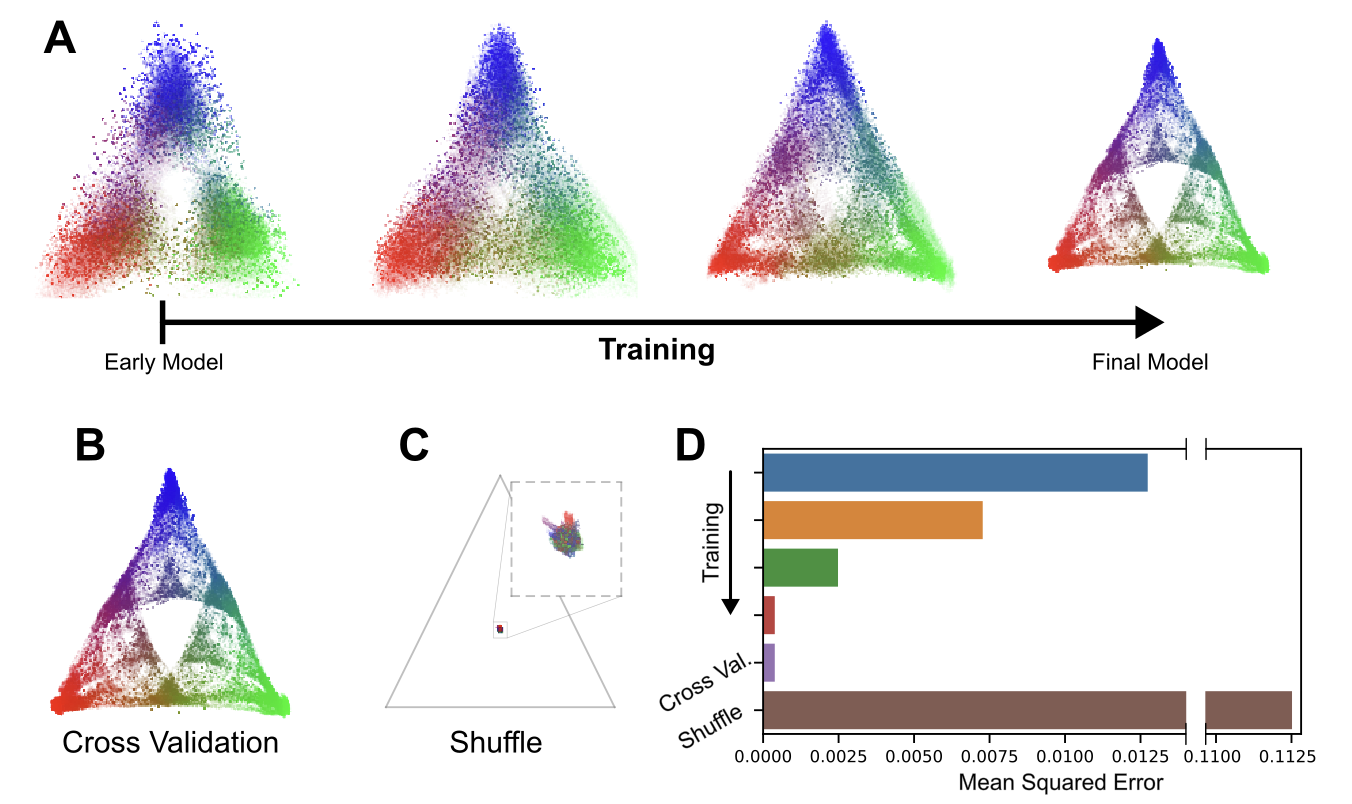}
    \caption{The representation of belief state geometry is nontrivial. (A) Projected activations at different stages of training shows the emergence of belief state geometry.  (B) Cross-validation of our main result. (C) We shuffle the belief states in our linear regression procedure, preserving the overall ground-truth fractal shape while getting rid of the associated context.  The new projection collapses the data, showing that the fractal's appearance in the residual stream is not an artifact of projecting high-dimensional data to a desired shape. (D) Mean squared error (averaged across input sequences) between (i) the position of projected activations and (ii) the ground-truth position of the corresponding belief state.}
    \label{fig:mess3_traiining_detail}
\end{figure}

We ran a number of controls, shown in Figure \ref{fig:mess3_traiining_detail}, to make sure these results were not artifacts. First, we performed our analysis at multiple points through training and found that the simplex structure emerged gradually (Figure \ref{fig:mess3_traiining_detail}A), suggesting that the detailed fractal structure found at the end of training was not a trivial consequence of the model architecture or initialization. Second, quantifying the mean squared error of the regression revealed a decrease in the errors of belief-state geometry representation throughout the course of training (Figure \ref{fig:mess3_traiining_detail}D, first 4 bars). In addition, more faithful representation of the belief state geometry in the residual stream corresponded to lower cross entropy loss (Figure \ref{fig:mse}). Third, we performed cross-validation, where only 20\% of all input-activation pairs were used to train the regression, and then the held out 80\% data was used to visualize and analyze the result, and found that the geometry of the belief state was still well represented (Figure \ref{fig:mess3_traiining_detail}B) and that the mean squared error was similar to that of the full regression (Figure \ref{fig:mess3_traiining_detail}D, red vs.\ purple bars). 
Finally, we ran a control that preserved the fractal geometry in the simplex but shuffled the input-point correspondences, resulting in the regression mapping all points to the simplex center due to the lack of discoverable structure (Figure \ref{fig:mess3_traiining_detail}C).

These results provide compelling evidence for our central claim: transformers trained on data with hidden generative structure will learn to represent the geometry of belief states in their residual stream. The close match between the predicted fractal structure and the empirical geometry of the residual activations, even for the highly complex Mess3 MSP, suggests that this geometry is a fundamental aspect of how transformers build predictive representations of their input data.

\subsection{Belief state geometry represents information beyond next-token prediction}


Often, \emph{distinct} belief states will have the \emph{same} next-token prediction associated with them. Our framework suggests that  transformers will keep internal distinctions in the representations of these belief states, despite the fact that transformers are trained explicitly on next-token prediction.

\begin{figure} [t]
    \centering
    \includegraphics[width=0.75\linewidth]{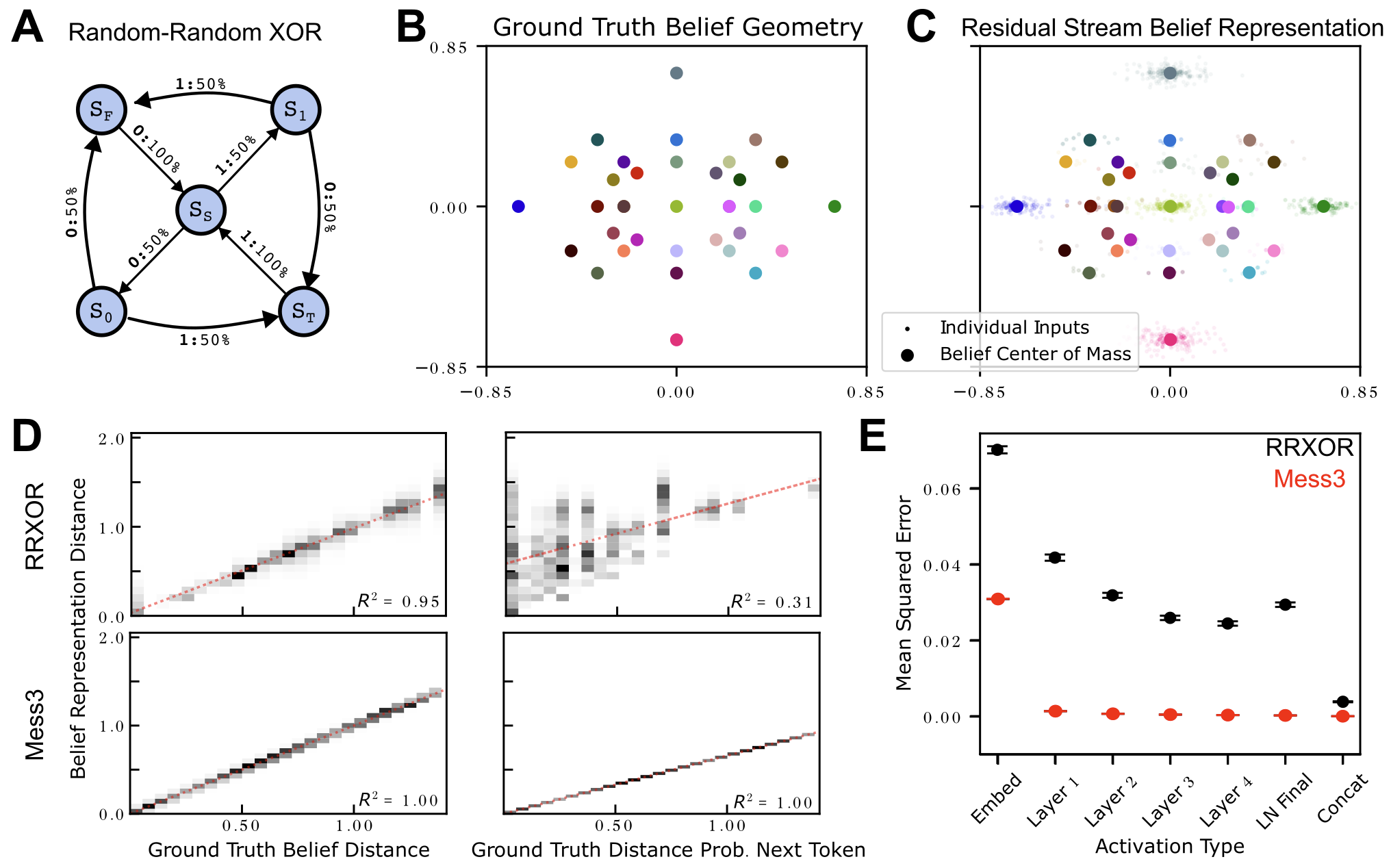}
    \caption{The belief state geometry can be represented across multiple layers when the belief state structure has next-token degeneracies. (A) The Random-Random-$\texttt{XOR}$ process has zero pairwise correlation, but has interesting higher-order structure. The MSP of this process has 36 distinct states, with many of them degenerate in terms of their optimal next-token predictions. (B) The belief state geometry lies in a 4-simplex. To visualize we project down to 2-dimensions, with each belief state colored uniquely. (C) The transformer linearly represents this belief state geometry. Small dots correspond to individual activations and are colored according to the ground truth belief state. Large dots correspond to the center of mass of all activations associated with a particular ground-truth belief state. (D)We compare Euclidean distances between pairs of ground truth belief states and see if those distances are preserved in the belief state representation in the transformer (left scatter plots), showing good correspondence for both RRXOR and Mess3 processes. The right scatter plots compare distances in ground truth next-token probabilities to distances in the transformer's belief state representations, showing low correlation for RRXOR ($R^2=0.31$) and high correlation for Mess3, indicating that RRXOR belief state geometry is not captured by next-token predictions alone.
   (E) 
   Mean squared error of the linear regression procedure, applied to individual layers and, at the far right, applied to the concatenation of activations across layers.  
   }
    \label{fig:RRX}
\end{figure}

The Random--Random--\texttt{XOR} (RRXOR) process, shown in Figure~\ref{fig:RRX}A,  has these types of degeneracies~\cite{Riec18_SSAC2}. The MSP of the RRXOR process has 36 distinct belief states, and can be geometrically represented in a 4-simplex. In Figure \ref{fig:RRX}B we visualize the simplex geometry by projecting down to 2 dimensions. After training, we run our regression analysis to see if the belief state geometry is represented in the residual stream. Unlike our previous results for the Mess3 process, we find that the belief state geometry is not represented in the final residual stream before the unembedding. However, when we take the residual stream activations of all layers and concatenate them, we find a veridical representation (Figure \ref{fig:RRX}C).

The geometry we find is not explainable by next-token predictions. To quantify this, we ask if the pairwise distances between belief state representations in the transformer can be explained by pairwise distances in the next-token predictions, or if they are better explained by ground truth belief state distances. First, we compute the Euclidean distance between pairs of ground truth belief states. For each pair, we also compute distances between the corresponding belief state representations in the transformer. In (Figure \ref{fig:RRX}D Left) we compare the ground truth pairwise distances to the represented pairwise distances, and find that for both the RRXOR and Mess3 process, there is good correspondence.

Next, for every pairwise distances of belief state representations, we compute the corresponding distance in the ground truth next token probabilities. The scatter plots on the right of Figure~\ref{fig:RRX}D compares these distances. For the RRXOR process, the correlation with next-token predictions is relatively low ($R^2 = 0.31$) compared to the correlation with the belief state geometry ($R^2 = 0.95$), indicating that belief state geometry captures structure in the residual stream not captured by next-token predictions. For the Mess3 process, the pairwise structure of next-token predictions is preserved in the belief state representation. This difference between the RRXOR and Mess3 results is expected from the theory. Since the RRXOR process contains distinct belief states with the same optimal next-token distribution, the discovered belief state geometry is not well-explained by the next token predictions.

\subsection{Belief state geometry can be spread across layers of the residual stream}

The framework presented here suggests that belief states should be represented in transformers, but does not say that they must be represented in the final layer. In cases of belief states that are degenerate in their next-token predictions, distinctions can be lost before the unembedding. Thus, in the Mess3 process we expect belief state geometry to persist to the final layer of the residual stream, whereas in the RRXOR trained transformer belief state information can be collapsed before the unembedding. 

To quantify this, we report the mean squared error of the belief geometry regression over the different layers of the transformer as well as the concatenation of residual activations from all layers in Figure~\ref{fig:RRX}E. We find that the RRXOR belief state geometry is not well represented in the final layer of the transformer.
In contrast, the  Mess3 belief state geometry is well represented in indidividual layers of the transformer, and persists through to the final layer before the unembedding. 
Interestingly, the RRXOR belief state geometry is not represented in \emph{any} of the individual layers of the transformer, but is represented in the concatenation of the layers. This suggests that the belief state geometry is spread across multiple layers of the residual stream.


\section{Discussion}
\label{sec:Discussion}


\subsection{Belief state geometry: Why? }

During training, models do not directly see the hidden structure generating the data---they only see data.  So how is it that the pretrained model infers belief states in the simplex of the generator states?  In fact, there are infinitely many distinct HMMs that can generate the same stochastic process~\citep{Crut10a}. Despite this, any stochastic process has a canonical vector-space representation in the space of probability densities over all possible future token sequences~\citep{Uppe97a}. 
This corresponds to a minimal HMM representation, and we should expect 
that transformers learn 
belief state geometry in the simplex over these minimal generator states. This canonical geometry for a process provides an explanation for why we were able to find the belief state geometry represented in our transformers.

In the jargon of computational mechanics, it is notable that the recurrent belief states map onto the causal states 
of the process' $\epsilon$-machine~\citep{Crut12a}.  The recurrent states must be distinguished for prediction, even if there exists a generative HMM that is smaller than the $\epsilon$-machine.
Typically, data can be generated by a much smaller mechanism than is required to predict it~\cite{lohr2012predictive, ruebeck2018prediction}. 
Accordingly, transformers and other pretrained models learn much more than just an accurate generative model of the world---they also learn how to synchronize to the hidden state of the world through observed context.



We expect our main results do not depend strongly on the particular neural network architecture of transformers, except that they (i) utilize a residual stream and (ii) were pretrained on future-token prediction.
Indeed, we expect to find the same belief state geometry in other neural network architectures like Mamba~\citep{Gu23_Mamba}, and recent work building on our discovery has shown that recurrent neural networks do represent the belief state geometry for the processes tested in our experiments \cite{rnn_belief}.
The generality of our results is thus a powerful architecture-independent insight for interpretability.  Pretraining will generally induce mixed-state geometry.  We should then be able to work backwards from this knowledge and the knowledge of the neural network architecture to determine both its world model and how it performs Bayesian updating over its hidden states.


Corollary 2 of \cite{Shal98a} implies that, for a recurrent neural network to perform as well as possible on next token prediction, all information about the past necessary to predict the 
entire future as well as possible 
must be present in the latent state of the network.  The implication for feedforward networks like transformer architectures is much less obvious, but it 
motivates the hypothesis
that
to perform as well as possible on next-token prediction requires that all information about the past necessary to predict the entire future as well as possible must be present across the layers of each context window position's residual stream.  Our results here support this interpretation.

Intuitively, why would a model learn about the entire future if it is only trained on next-token prediction?
The simple answer is that, even if different belief states imply the same probability distribution over the next token, 
any distinction in probability distributions over the entire future
may \emph{eventually}, after further context, imply distinct distributions over the next token at some point in the future.  If the initial nuance were discarded, then the ability to distinguish next-token probabilities in the future would be compromised.
To minimize loss, pretrained models need to not only nail the next-token probability distribution, but also retain all information necessary to get the correct next-token distributions in the distant future.  It turns out that this requires distinguishing all pasts that induce distinct probability distributions over the entire future~\citep{Shal98a}.

It is tempting to think that all of this information will persist at the final layer of the residual stream, but this is not strictly necessary nor borne out in practice as we showed in Figure~\ref{fig:RRX}.  In general, the belief state is spread across the layers at each position.  This is consistent with 
\cite{Pal23_Future},
which found that the 
predictive accuracy
of the residual stream with respect to far-future tokens peaks somewhere in intermediate layers.
Performing as well as possible on multi-token prediction, as in 
\cite{Gloeckle24_Better}, 
would imply the same belief-state structure as next-token prediction, 
but it should change how this information is spread across the layers and, in practice, performance may be different when loss is not minimized.

It is also interesting to consider the linearity of the belief state representations we found, and if the linearlity is strictly necessary for a network to optimize next-token cross-entropy loss. In fact, recent work has constructed a transformer by hand (i.e. a human being selected the weights) that perfectly solves the RRXOR task, but where the belief state geometry \textit{is not} linearly represented~\citep{GoldsteinHand}. It is thus a non-trivial empirical finding that transformer architectures trained via stochastic gradient descent find solutions in which the belief state geometry \textit{is} linearly represented.

\subsection{Further implications}

Our finding of the belief-state simplex in the residual stream strongly suggests 
that any model capable of minimal loss requires as many residual-stream dimensions 
as the number of states in a minimal generative model of the stochastic process sampled by the training data.
Moreover, we should be able to use this new understanding to determine the minimal loss with fewer residual-stream dimensions.  Likewise, using an adaptation of the rate distortion theory applied to the belief-state simplex~\citep{Marz17a}, we anticipate that our framework should be able to make non-trivial predictions about the evolving geometry of activations during training.

To the extent that different users are represented in the training data, they can be thought of as disconnected ergodic components of a non-ergodic stochastic process.  Even after the model has learned, it needs to synchronize (in context) to both (i) the ergodic component representing the user and (ii) the current latent state of that component.  Each will show up in different ways in the MSP.  
Using this framework, we should be able to predict rates of adaptations to different users---i.e., rates of convergence in context---as identified by the reduction in next-token entropy as context position increases.

\subsection{What are features?}
A growing literature in LLM research studies what \emph{features} these systems represent~\cite{bricken2023monosemanticity}. Importantly, there is no currently agreed upon definition of feature. Consequently, it has been difficult to study this issue in a principled manner in which ground truth features are known. While the MSP provides a ground truth understanding of the computation next-token predictors are asked to accomplish, the relationship between belief states and features is not necessarily one-to-one. A single belief state may encompass multiple features, and the same feature may be represented across multiple belief states. The exact mapping between belief states and features is likely to be complex and dependent on the specific architecture and training data of the transformer model, making it an exciting open problem for future research.

\subsection{Limitations and applicability to more realistic settings}

Our experimental validation focused on small-scale systems, using HMMs with only 3-5 states and vocabularies of 2-3 tokens. While these systems exhibited meaningful complexity through infinite Markov order in both the RRXOR and Mess3 processes, they represent a significant simplification compared to the sophisticated architectures and massive vocabularies (>50,000 tokens) of modern language models trained on natural language data. This scale limitation raises important questions about how our framework extends to larger and more complex settings.

For realistic systems like board games or natural language, the dimensionality of the belief state simplex would exceed the dimension of the residual stream. This suggests that if transformers represent belief state geometry in these domains, they must do so in a compressed form. Understanding the nature of this compression—what information is preserved and what is discarded—represents a crucial direction for future work.

As the size of the minimal generating structure and the context window length grows, explicitly computing ground-truth belief states becomes intractable. We will need new methodologies for validating our framework in settings where we cannot directly compute the MSP structure.

We limited ourselves to the simpler stationary ergodic setting in this paper to empirically verify important basic features of transformer representations. However, the framework presented in this work will naturally extend to non-stationary~\cite{riechers2021fraudulent} and non-ergodic processes~\cite{crutchfield2015signatures}, which is important for the extension to real-world tasks. For example, in in-context learning the underlying process is inherently nonergodic.  In the more general setting, we also expect transformers to represent belief-state geometry, although that geometry will then reflect the more general types of non-stationary and non-ergodic processes representative of real-world data like natural language~\cite{debowski2020information, dkebowski2023simplistic}.

In the experiments we ran, we generated training data using HMMs. A natural question is what to expect with training data generated from non-HMM sources. It is important to note that any dataset made of sequences of tokens used to train a transformer can be represented as being generated from an HMM, if one allows for non-ergodicity, non-stationarity, and infinite states. For example, in the case of training data that takes the form of parenthesis matched strings, one would usually conceive of this as being generated by a push-down automaton. However, this can equivalentaly be represented by an HMM consisting of an infinite chain of hidden states. Another interesting point that we did not focus on in our presentation for the sake of simplicity, is that even if one does not assume an HMM for the generating machine, the question of the computational structure of an optimal predictor is naturally an HMM, with hidden states given by distinct distributions over the future (whether the generator is an HMM or a Turing Machine, or something else)~\cite{Uppe97a}.




In this work we verified that belief states are linearly represented in the residual stream, even in cases where the belief states form intricate fractal geometries. Although this is quite suggestive that these representations are used to implement some form of Bayesian updating, we did not directly test this. In other words, we do not yet know if these LLMs develop something like a circuit for implementing Bayesian updates, as would be natural for MSP dynamics. So far, we have only established
the representation of belief states and their geometry.

Computational mechanics is primarily (but not completely, see Ref.~\cite{marzen2017nearly})  concerned with optimal prediction, but LLMs in practice are not perfectly optimal. Further work is needed to understand how near-optimality and non-optimality influence belief state representations. 
Progress in this direction 
could offer insights into 
evolving structures during
training.

Moving beyond stationary ergodic HMM training data, we still generically expect fractal-like structure in the residual-stream activations. Why? The fractal structure is a result of Bayesian updating applied to the “non-deterministic” computational structure of HMMs~\cite{jurgens2021divergent}. With stationary processes, this same Bayesian map is folded into itself time and again, producing a very clean fractal. For natural language, even in the absence of stationarity, there will nevertheless be resonances of non-deterministic computational structure, which would imply a type of folding beliefs in a self-similar manner. However, a more rigorous version of these statements and the empirical validation is left for future work.

\section{Conclusion}

In this work, we introduced a theoretical framework that establishes a clear connection between the structure of training data and the geometric properties of activations in trained transformer neural networks. 
Through experiments, we validated that the geometry of belief states is linearly represented within the residual stream of the transformer architecture. This finding suggests that transformers construct predictive representations that surpass simple next-token prediction. Instead, these representations encode the complex geometry associated with belief updating over hidden states, capturing an inference process over the underlying structure of the data-generating process.

Our work takes a significant step towards concretizing the understanding of the computational structures that drive the behavior of large language models (LLMs) and how these structures relate to the data on which they are trained. This concretization is crucial, as the rapid advancements in LLMs have led to models with increasingly sophisticated behavioral capabilities. However, without a clear understanding of the underlying computational structures and their relationship to the training data, it becomes challenging to interpret, trust, and further improve these models.


\medskip

\bibliographystyle{plainnat}
\bibliography{chaos,ref}

\begin{thebibliography}{33}
\providecommand{\natexlab}[1]{#1}
\providecommand{\url}[1]{\texttt{#1}}
\expandafter\ifx\csname urlstyle\endcsname\relax
  \providecommand{\doi}[1]{doi: #1}\else
  \providecommand{\doi}{doi: \begingroup \urlstyle{rm}\Url}\fi

\bibitem[Bachmann and Nagarajan(2024)]{bachmann2024pitfalls}
Gregor Bachmann and Vaishnavh Nagarajan.
\newblock The pitfalls of next-token prediction.
\newblock \emph{arXiv preprint arXiv:2403.06963}, 2024.

\bibitem[Bender et~al.(2021)Bender, Gebru, McMillan-Major, and Shmitchell]{bender2021dangers}
Emily~M Bender, Timnit Gebru, Angelina McMillan-Major, and Shmargaret Shmitchell.
\newblock On the dangers of stochastic parrots: Can language models be too big?
\newblock In \emph{Proceedings of the 2021 ACM conference on fairness, accountability, and transparency}, pages 610--623, 2021.

\bibitem[Bricken et~al.(2023)Bricken, Templeton, Batson, Chen, Jermyn, Conerly, Turner, Anil, Denison, Askell, Lasenby, Wu, Kravec, Schiefer, Maxwell, Joseph, Hatfield-Dodds, Tamkin, Nguyen, McLean, Burke, Hume, Carter, Henighan, and Olah]{bricken2023monosemanticity}
Trenton Bricken, Adly Templeton, Joshua Batson, Brian Chen, Adam Jermyn, Tom Conerly, Nick Turner, Cem Anil, Carson Denison, Amanda Askell, Robert Lasenby, Yifan Wu, Shauna Kravec, Nicholas Schiefer, Tim Maxwell, Nicholas Joseph, Zac Hatfield-Dodds, Alex Tamkin, Karina Nguyen, Brayden McLean, Josiah~E Burke, Tristan Hume, Shan Carter, Tom Henighan, and Christopher Olah.
\newblock Towards monosemanticity: Decomposing language models with dictionary learning.
\newblock \emph{Transformer Circuits Thread}, 2023.
\newblock https://transformer-circuits.pub/2023/monosemantic-features/index.html.

\bibitem[Crutchfield(2012)]{Crut12a}
J.~P. Crutchfield.
\newblock Between order and chaos.
\newblock \emph{Nature Physics}, 8\penalty0 (January):\penalty0 17--24, 2012.
\newblock \doi{10.1038/NPHYS2190}.

\bibitem[Crutchfield et~al.(2010)Crutchfield, Ellison, Mahoney, and James]{Crut10a}
J.~P. Crutchfield, C.~J. Ellison, J.~R. Mahoney, and R.~G. James.
\newblock Synchronization and control in intrinsic and designed computation: {An} information-theoretic analysis of competing models of stochastic computation.
\newblock \emph{CHAOS}, 20\penalty0 (3):\penalty0 037105, 2010.
\newblock Santa Fe Institute Working Paper 10-08-015; arxiv.org:1007.5354 [cond-mat.stat-mech].

\bibitem[Crutchfield and Marzen(2015)]{crutchfield2015signatures}
James~P Crutchfield and Sarah Marzen.
\newblock Signatures of infinity: Nonergodicity and resource scaling in prediction, complexity, and learning.
\newblock \emph{Physical Review E}, 91\penalty0 (5):\penalty0 050106, 2015.

\bibitem[Debowski(2020)]{debowski2020information}
Lukasz Debowski.
\newblock \emph{Information theory meets power laws: Stochastic processes and language models}.
\newblock John Wiley \& Sons, 2020.

\bibitem[D{\k{e}}bowski(2023)]{dkebowski2023simplistic}
{\L}ukasz D{\k{e}}bowski.
\newblock A simplistic model of neural scaling laws: Multiperiodic santa fe processes.
\newblock \emph{arXiv preprint arXiv:2302.09049}, 2023.

\bibitem[Engels et~al.(2024)Engels, Liao, Michaud, Gurnee, and Tegmark]{engels2024not}
Joshua Engels, Isaac Liao, Eric~J Michaud, Wes Gurnee, and Max Tegmark.
\newblock Not all language model features are linear.
\newblock \emph{arXiv preprint arXiv:2405.14860}, 2024.

\bibitem[Gloeckle et~al.(2024)Gloeckle, Idrissi, Rozi{\`e}re, Lopez-Paz, and Synnaeve]{Gloeckle24_Better}
Fabian Gloeckle, Badr~Youbi Idrissi, Baptiste Rozi{\`e}re, David Lopez-Paz, and Gabriel Synnaeve.
\newblock Better \& faster large language models via multi-token prediction.
\newblock \emph{arXiv preprint arXiv:2404.19737}, 2024.

\bibitem[Goldstein(2024)]{GoldsteinHand}
Rick Goldstein.
\newblock Handcrafting a network to predict next token probabilities for the random-random-xor process.
\newblock https://apartresearch.com/event/compmech, June 2024.
\newblock Research submission to the Computational Mechanics Hackathon! research sprint hosted by Apart, PIBBSS, and Simplex.

\bibitem[Gu and Dao(2023)]{Gu23_Mamba}
Albert Gu and Tri Dao.
\newblock Mamba: Linear-time sequence modeling with selective state spaces.
\newblock \emph{arXiv preprint arXiv:2312.00752}, 2023.

\bibitem[Ha and Schmidhuber(2018)]{ha2018world}
David Ha and J{\"u}rgen Schmidhuber.
\newblock World models.
\newblock \emph{arXiv preprint arXiv:1803.10122}, 2018.

\bibitem[Hu et~al.(2024)Hu, Liu, and Jin]{hu2024limitation}
Jiachen Hu, Qinghua Liu, and Chi Jin.
\newblock On limitation of transformer for learning hmms.
\newblock \emph{arXiv preprint arXiv:2406.04089}, 2024.

\bibitem[Jurgens and Crutchfield(2021{\natexlab{a}})]{Jurg21_Shannon}
A.~M. Jurgens and J.~P. Crutchfield.
\newblock Shannon entropy rate of hidden {M}arkov processes.
\newblock \emph{Journal of Statistical Physics}, 183\penalty0 (2):\penalty0 32, 2021{\natexlab{a}}.

\bibitem[Jurgens and Crutchfield(2021{\natexlab{b}})]{jurgens2021divergent}
Alexandra~M Jurgens and James~P Crutchfield.
\newblock Divergent predictive states: The statistical complexity dimension of stationary, ergodic hidden markov processes.
\newblock \emph{Chaos: An Interdisciplinary Journal of Nonlinear Science}, 31\penalty0 (8), 2021{\natexlab{b}}.

\bibitem[Li et~al.(2022)Li, Hopkins, Bau, Vi{\'e}gas, Pfister, and Wattenberg]{li2022emergent}
Kenneth Li, Aspen~K Hopkins, David Bau, Fernanda Vi{\'e}gas, Hanspeter Pfister, and Martin Wattenberg.
\newblock Emergent world representations: Exploring a sequence model trained on a synthetic task.
\newblock \emph{arXiv preprint arXiv:2210.13382}, 2022.

\bibitem[Liu et~al.(2022)Liu, Ash, Goel, Krishnamurthy, and Zhang]{liu2022transformers}
Bingbin Liu, Jordan~T Ash, Surbhi Goel, Akshay Krishnamurthy, and Cyril Zhang.
\newblock Transformers learn shortcuts to automata.
\newblock \emph{arXiv preprint arXiv:2210.10749}, 2022.

\bibitem[L{\"o}hr(2012)]{lohr2012predictive}
Wolfgang L{\"o}hr.
\newblock Predictive models and generative complexity.
\newblock \emph{Journal of Systems Science and Complexity}, 25:\penalty0 30--45, 2012.

\bibitem[Marzen and Crutchfield(2017{\natexlab{a}})]{Marz17a}
S.~E. Marzen and J.~P. Crutchfield.
\newblock Nearly maximally predictive features and their dimensions.
\newblock \emph{Phys. Rev. E}, 95\penalty0 (5):\penalty0 051301(R), 2017{\natexlab{a}}.
\newblock \doi{10.1103/PhysRevE.95.051301}.
\newblock SFI Working Paper 17-02-007; arxiv.org:1702.08565 [cond-mat.stat-mech].

\bibitem[Marzen and Crutchfield(2017{\natexlab{b}})]{marzen2017nearly}
Sarah~E Marzen and James~P Crutchfield.
\newblock Nearly maximally predictive features and their dimensions.
\newblock \emph{Physical Review E}, 95\penalty0 (5):\penalty0 051301, 2017{\natexlab{b}}.

\bibitem[Nanda and Bloom(2022)]{Nanda2022_Transformerlens}
Neel Nanda and Joseph Bloom.
\newblock Transformerlens.
\newblock \url{https://github.com/TransformerLensOrg/TransformerLens}, 2022.

\bibitem[Nanda et~al.(2023)Nanda, Lee, and Wattenberg]{nanda2023emergent}
Neel Nanda, Andrew Lee, and Martin Wattenberg.
\newblock Emergent linear representations in world models of self-supervised sequence models.
\newblock \emph{arXiv preprint arXiv:2309.00941}, 2023.

\bibitem[Ortega et~al.(2019)Ortega, Wang, Rowland, Genewein, Kurth-Nelson, Pascanu, Heess, Veness, Pritzel, Sprechmann, et~al.]{ortega2019meta}
Pedro~A Ortega, Jane~X Wang, Mark Rowland, Tim Genewein, Zeb Kurth-Nelson, Razvan Pascanu, Nicolas Heess, Joel Veness, Alex Pritzel, Pablo Sprechmann, et~al.
\newblock Meta-learning of sequential strategies.
\newblock \emph{arXiv preprint arXiv:1905.03030}, 2019.

\bibitem[Pal et~al.(2023)Pal, Sun, Yuan, Wallace, and Bau]{Pal23_Future}
Koyena Pal, Jiuding Sun, Andrew Yuan, Byron~C Wallace, and David Bau.
\newblock Future lens: Anticipating subsequent tokens from a single hidden state.
\newblock \emph{arXiv preprint arXiv:2311.04897}, 2023.

\bibitem[Pepper(2024)]{rnn_belief}
Keenan Pepper.
\newblock {RNNs} represent belief state geometry in their hidden states.
\newblock https://apartresearch.com/event/compmech, June 2024.
\newblock Research submission to the Computational Mechanics Hackathon! research sprint hosted by Apart, PIBBSS, and Simplex.

\bibitem[Riechers and Crutchfield(2018{\natexlab{a}})]{Riec18_SSAC1}
P.~M. Riechers and J.~P. Crutchfield.
\newblock Spectral simplicity of apparent complexity, {Part I}: {The} nondiagonalizable metadynamics of prediction.
\newblock \emph{Chaos}, 28:\penalty0 033115, 2018{\natexlab{a}}.
\newblock \doi{10.1063/1.4985199}.

\bibitem[Riechers and Crutchfield(2018{\natexlab{b}})]{Riec18_SSAC2}
P.~M. Riechers and J.~P. Crutchfield.
\newblock Spectral simplicity of apparent complexity, {Part II}: {Exact} complexities and complexity spectra.
\newblock \emph{Chaos}, 28:\penalty0 033116, 2018{\natexlab{b}}.
\newblock \doi{10.1063/1.4986248}.

\bibitem[Riechers and Crutchfield(2021)]{riechers2021fraudulent}
Paul~M Riechers and James~P Crutchfield.
\newblock Fraudulent white noise: Flat power spectra belie arbitrarily complex processes.
\newblock \emph{Physical Review Research}, 3\penalty0 (1):\penalty0 013170, 2021.

\bibitem[Ruebeck et~al.(2018)Ruebeck, James, Mahoney, and Crutchfield]{ruebeck2018prediction}
Joshua~B Ruebeck, Ryan~G James, John~R Mahoney, and James~P Crutchfield.
\newblock Prediction and generation of binary markov processes: Can a finite-state fox catch a markov mouse?
\newblock \emph{Chaos: An Interdisciplinary Journal of Nonlinear Science}, 28\penalty0 (1), 2018.

\bibitem[Shalizi and Crutchfield(2001)]{Shal98a}
C.~R. Shalizi and J.~P. Crutchfield.
\newblock Computational mechanics: Pattern and prediction, structure and simplicity.
\newblock \emph{J. Stat. Phys.}, 104:\penalty0 817--879, 2001.

\bibitem[Sutskever(2023)]{Sutskever2023}
Ilya Sutskever.
\newblock Building {AGI}, alignment, future models, spies, {M}icrosoft, {T}aiwan, \& enlightenment.
\newblock Podcast on Dwarkesh Patel Podcast, March 2023.
\newblock URL \url{https://www.youtube.com/watch?v=YEUclZdj_Sc}.
\newblock Accessed: 2024-05-22.

\bibitem[Upper(1997)]{Uppe97a}
D.~R. Upper.
\newblock \emph{Theory and Algorithms for Hidden {M}arkov Models and Generalized Hidden {M}arkov Models}.
\newblock PhD thesis, University of California, Berkeley, 1997.
\newblock {P}ublished by University Microfilms Intl, Ann Arbor, Michigan.

\end{thebibliography}

%




\appendix

\section{Appendix / supplemental material}

\subsection{Supplemental Figures}
\newcounter{supfigure}
\renewcommand{\thesupfigure}{S\arabic{supfigure}}

\newenvironment{suppfigure}{
     \refstepcounter{supfigure} 
     \renewcommand\thefigure{\thesupfigure}%
     \addtocounter{figure}{-1}%
     \begin{figure}}
    {\end{figure}}

\begin{suppfigure}[H]
    \centering
    \includegraphics[width=0.75\linewidth]{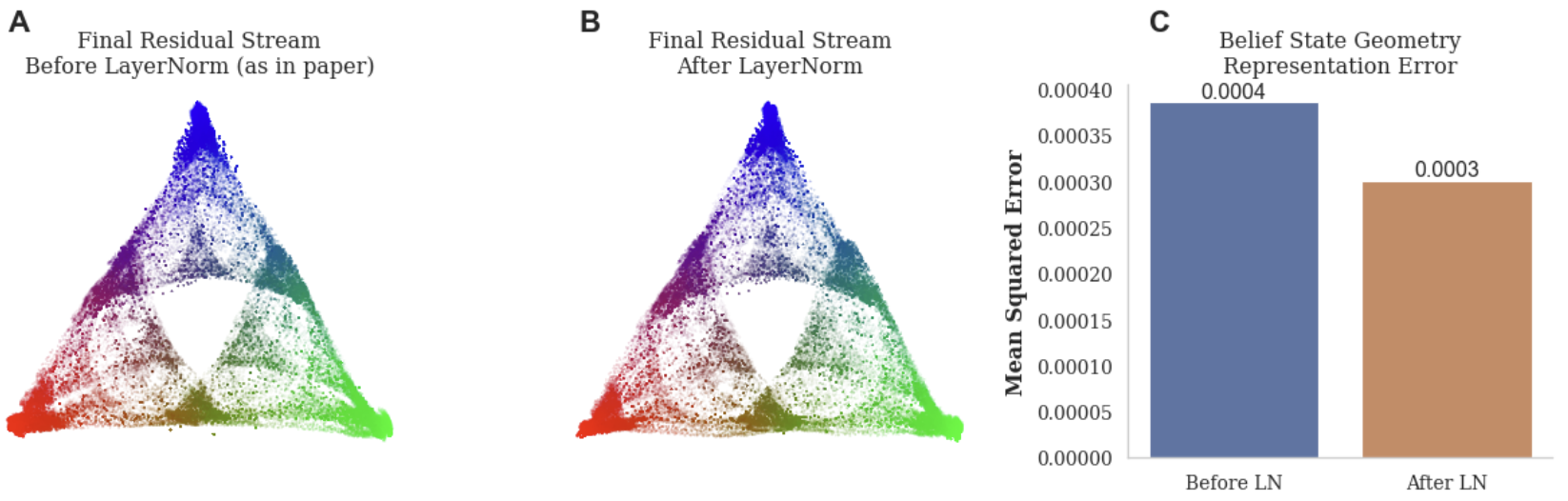}
    \caption{To test the effect of using residual stream activations from before or after the final LayerNorm in our analysis, we compared the belief state geometry representations in both cases. (A) Projection of the final residual stream activations before LayerNorm onto the belief state simplex, as presented in the main paper. (B) The same projection for activations after LayerNorm, showing a qualitatively similar structure. (C) Mean squared error of the linear fit capturing the belief state geometry for both cases. The representation after LayerNorm shows a slightly lower error (0.0003) compared to before LayerNorm (0.0004), indicating that the belief state geometry is preserved and marginally better represented after the LayerNorm operation. These results demonstrate that our findings are robust to the choice of using pre- or post-LayerNorm activations.}
    \label{fig:layernorm}
\end{suppfigure}

\begin{suppfigure}[H]
    \centering
    \includegraphics[width=0.75\linewidth]{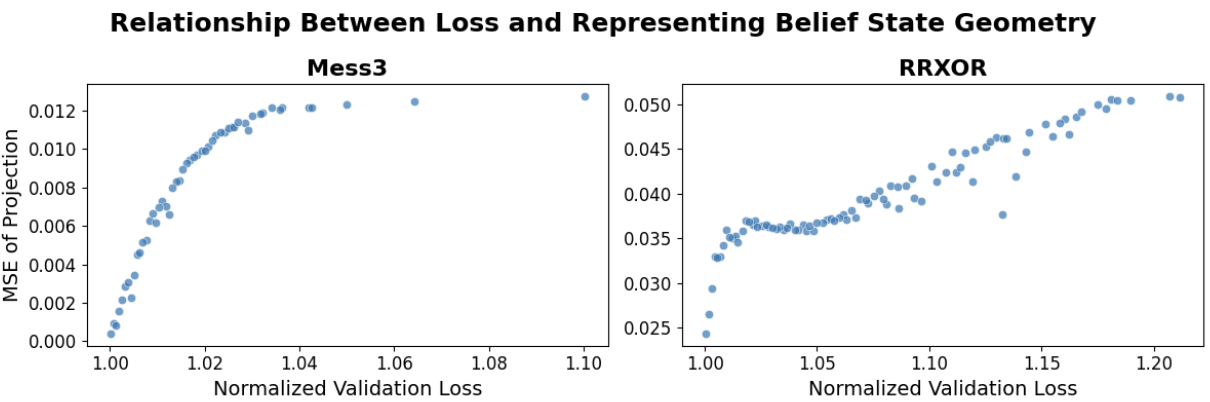}
    \caption{To quantify the relationship between transformer performance on next-token prediction and representing the belief state geometry in the residual stream, we plotted normalized validation loss vs. mean squared error (MSE) of projecting residual stream activations onto the belief state geometry for Mess3 (left) and RRXOR (right) processes. Validation loss is normalized to the theoretical optimal, such that 1.0 characterizes a transformer with optimal loss. As the validation loss decreases (moving left on the x-axis), the MSE of the regression also decreases, indicating that better performance on the next-token prediction task correlates with a more accurate representation of the belief state geometry in the residual stream.}
    \label{fig:mse}
\end{suppfigure}

\subsection{Relations to Previous Work}

A number of connections to previous work in computational mechanics have been mentioned in the Discussion section. Here we will continue the discussion of related works, focusing on connections to neural network research.

\paragraph{Transformer Internal Representations}
Recent work has significantly advanced our understanding of how and what information is encoded within transformer models. Ref.~\citep{Pal23_Future} introduced the "Future Lens" framework, demonstrating that individual hidden states in transformers trained on natural language contain information about multiple future tokens. This finding is consistent with transformers representing belief states, since belief states contain information well beyond the next token. Ref.~\citep{engels2024not} found that the internal representations of days of the week in LLMs lie in particular geometric arrangements (in a circular pattern). Since our work draws a concrete relation between the structure of data and geometric arrangements inside of transformers, a promising avenue of research would be to explain the days of the week result using our framework.
Other work has studied internal representations in the board game Othello, finding a representation of the board state in the residual stream~\cite{li2022emergent, nanda2023emergent}. This finding is interpreted by the authors as the transformers containing a “world model”. Conceptually extending our work to such a setting is quite natural. In a sequential game with full knowledge, such as Tic-tac-toe, Othello, and Chess, each board state uniquely determines the future distribution of moves. Thus, the belief states correspond to board states, providing an explanation for why transformers trained on gameplay should contain explicit representations of board states in their residual stream. 
However, this also makes clear a limitation of our approach. The simplex associated with these belief states would, for any realistic board game, be of larger dimension than the residual stream. If transformers are representing the belief state geometry in those cases, they must be doing so in a compressed fashion. This is an important theoretical problem for future work. Importantly, our work provides a unique model-agnostic point of view to study interpretability.
\paragraph{Behavioral Limitations of Transformers and Chain of Thought}
A number of papers have studied the limitations of transformers to learn certain algorithmic tasks, such as graph path-finding tasks and HMMs~\cite{bachmann2024pitfalls, hu2024limitation}. These papers provide important empirical results that are interesting to think through the perspective of the MSP and belief state geometry. 
In particular, Ref.~\cite{hu2024limitation} provides an upper-bound for the depth of a transformer needed to capture HMM generated data up to a particular context-window length. This makes sense given that transformers are feedforward machines, and the MSP is, in general, an automata containing recurrent components. Thus, what the transformer should be capturing is really an unrolled version of the MSP, which naturally gives an upper bound for how many layers one would need to capture the MSP to a certain depth. This paper also explicitly performs experiments on belief-state inference. In some of their experiments the explicitly train transformers to map sequences of emissions from an HMM to sequences of the associated belief states. In other experiments they train in the normal autoregressive manner solely on the observations, like in our study. Our work provides a nice theoretical connection between the two tasks.

In Ref.~\citep{bachmann2024pitfalls}, the authors test and explain a lack of ability for transformers to perform seemingly simple graph planning tasks. In this work, and others like it that study particular algorithmic tasks with particular formal structures like finite-automata~\cite{liu2022transformers}, addition, and the like, we think its interesting to think of how our framework would apply. In general, our approach treats the sequential training data as a first class citizen. This makes clear that to relate studies of formal structures (like graph planning) to our framework, one must explicitly consider the particular tokenization used to convert from the formal structure to sequences of tokens. Then the main consideration becomes the computational structure of the sequential nature of the tokens generated by the chosen tokenization scheme (as opposed to the formal structure directly). From our perspective what one learns, and the limitations of being able to perform well on next-token generation given that tokenization, is directly related to the tokenization and not directly to the formal structure that was tokenized. Thus, these results provide important empirical results for our framework to explain, or good tests to find where our framework fails and needs to be amended/expanded.

\paragraph{Meta-learning}
In Ref.~\cite{ortega2019meta}, the authors provide a theoretical framework for predictors and agents based on a Bayesian framework which builds state machines of sufficient statistics. This is analogous to the mixed-state presentation we have presented here. Indeed, work in computational mechanics has shown that the states of the MSP correspond not only to beliefs over the generator states, but also to the minimal sufficient statistics for predicting the future based on the past. One exciting area of further research this exposes is the potential for computational mechanics to inform our understanding of artificial RL agents.

\subsection{Details of data generating processes used in experiments}
The transition matrices for the Mess3 process \(T^{(A)}\), \(T^{(B)}\), and \(T^{(C)}\) are defined as follows:
\begin{align*}
T^{(A)} &= 
\begin{pmatrix}
0.765 & 0.00375 & 0.00375 \\
0.0425 & 0.0675 & 0.00375 \\
0.0425 & 0.00375 & 0.0675
\end{pmatrix},
\quad
T^{(B)} = 
\begin{pmatrix}
0.0675 & 0.0425 & 0.00375 \\
0.00375 & 0.765 & 0.00375 \\
0.00375 & 0.0425 & 0.0675
\end{pmatrix}, \text{ and}\\
T^{(C)} &= 
\begin{pmatrix}
0.0675 & 0.00375 & 0.0425 \\
0.00375 & 0.0675 & 0.0425 \\
0.00375 & 0.00375 & 0.765
\end{pmatrix} ~.
\end{align*}

The RRXOR process has transition matrices defined as 
\[
T^{(0)} = 
\begin{pmatrix}
0 & 0.5 & 0 & 0 & 0 \\
0 & 0 & 0 & 0 & 0.5 \\
0 & 0 & 0 & 0.5 & 0 \\
0 & 0 & 0 & 0 & 0 \\
1 & 0 & 0 & 0 & 0
\end{pmatrix} ~ \text{ and }
\quad
T^{(1)} = 
\begin{pmatrix}
0 & 0 &  0.5 & 0 & 0 \\
0 & 0 & 0 & .5 & 0 \\
0 & 0 & 0 & 0 & .5 \\
1 & 0 & 0 & 0 & 0 \\
0 & 0 & 0 & 0 & 0
\end{pmatrix} ~,
\]
where \( T^{(0)} \) and \( T^{(1)} \) are the transition matrices for emissions 0 and 1, respectively.

\subsection{Belief State Simplex}
Belief states
lie within the probability simplex \( \Delta^{|\mathcal{S}|-1} \), defined as:
\[
\Delta^{|\mathcal{S}|-1} = \Bigl\{ b \in \mathbb{R}^{|\mathcal{S}|} : \, \sum_{i=1}^{|\mathcal{S}|} b_i = 1 \text{ and } b_i \geq 0 \text{ for all } i \Bigr\}~.
\]

\subsection{Regression}
For the regression procedure used in this paper we used standard linear regression, assuming an affine model for the data and then minimizing:

\[
\min_{W, c} \sum_{i} \left\| b_i - (W a_i + c) \right\|^2
\]

where the summation is over all input sequences in the dataset.

\subsection{Details of transformer architecture and training}
In our experiments, we trained a transformer model using the following hyperparameters and training parameters: The model had a context window size of 10, used ReLU as the activation function, and had a head dimension of 8 and a model dimension of 64. There was 1 attention head in each of 4 layers. The model had MLPs of dimension 256 and used causal attention masking. Layer normalization was applied. For training, we used the Stochastic Gradient Descent (SGD) optimizer with a batch size of 64, running for 1,000,000 epochs, and a learning rate of 0.01 with no weight decay. For each batch we generated 64 sequences from the Mess3 or RRXOR HMM, choosing an initial hidden state from the stationary distribution. We instantiated the networks and performed our analysis using the TransformerLens library~\cite{Nanda2022_Transformerlens}.

\subsection{Details for analysis}
For the shuffle and cross validation analyses used in Figure \ref{fig:mess3_traiining_detail}, we performed both the shuffle and the cross validation procedure 1000 times over, randomly shuffling or performing the train test split independently 1000 times.

\subsection{Reproducibility Instructions}

To install the necessary dependencies and the code, follow these steps:

\subsubsection{Installation}

Navigate to the repository folder and run:
\begin{verbatim}
pip install -e .
\end{verbatim}

\subsubsection{Reproducing Figures}

To reproduce the figures presented in the paper, step through the provided Jupyter notebooks:

\begin{itemize}
    \item \texttt{Figure6.ipynb}
    \item \texttt{Figure7.ipynb}
\end{itemize}

These notebooks utilize the saved models located in the \texttt{/examples/models} directory.

\subsubsection{Training the Models}

The models were trained using legacy code found in the following scripts:

\begin{itemize}
    \item \texttt{legacy/epsilon-transformers/run\_sweeps\_mess3.py}
    \item \texttt{legacy/epsilon-transformers/run\_sweeps\_RRXOR.py}
\end{itemize}

These instructions should ensure that the environment is correctly set up and that the figures can be reproduced as described in the paper.



\end{document}